\documentclass{article}

\usepackage[preprint]{neurips_2023}
\usepackage[utf8]{inputenc}
\usepackage[T1]{fontenc}
\usepackage{hyperref}
\usepackage{url}
\usepackage{booktabs}
\usepackage{amsfonts}
\usepackage{amsmath}
\usepackage{amssymb}
\usepackage{nicefrac}
\usepackage{microtype}
\usepackage{xcolor}
\usepackage{graphicx}
\usepackage{subcaption}
\usepackage{algorithm}
\usepackage{algorithmic}
\usepackage{enumitem}
\usepackage{amsthm}
\usepackage{ragged2e}
\usepackage{placeins}
\usepackage{float}
\usepackage{fancyhdr}

\fancypagestyle{plain}{%
  \fancyhf{}
  
  \fancyfoot[C]{\footnotesize Preprint --- version 2 (January 2026) \hfill \thepage}
}
\fancypagestyle{empty}{%
  \fancyhf{}
  
  \fancyfoot[C]{\footnotesize Preprint --- version 2 (January 2026) \hfill \thepage}
}

\makeatletter
\renewcommand{\@noticestring}{}
\makeatother

\theoremstyle{definition}

\usepackage{titlesec}
\titlespacing*{\section}{0pt}{12pt}{6pt}
\titlespacing*{\subsection}{0pt}{10pt}{4pt}
\titlespacing*{\subsubsection}{0pt}{8pt}{3pt}
\titlespacing*{\paragraph}{0pt}{6pt}{0.5em}

\setlist[itemize]{leftmargin=*,noitemsep,topsep=3pt}

\title{%
  \textbf{Geometric Dynamics of Agentic Loops in Large Language Models}\\[0.4em]
  \large{Trajectories, Attractors and Dynamical Regimes in Semantic Space}
}

\author{
  Nicolas Tacheny \\
  University of Mons (Mons, Belgium) \\
  \texttt{nicolas.tacheny@gmail.com} \\
}

\begin{document}

\RaggedRight

\maketitle

\begin{abstract}
Iterative LLM systems---self-refinement, chain-of-thought, autonomous agents---are increasingly deployed, yet their temporal dynamics remain uncharacterized.
Prior work evaluates task performance at convergence but ignores the \emph{trajectory}: how does semantic content evolve across iterations? Does it stabilize, drift, or oscillate?
Without answering these questions, we cannot predict system behavior, guarantee stability, or systematically design iterative architectures.

\textbf{We formalize agentic loops as discrete dynamical systems in semantic space.}
Borrowing from dynamical systems theory, we define \emph{trajectories}, \emph{attractors} and \emph{dynamical regimes} for recursive LLM transformations, providing rigorous geometric definitions adapted to this setting.
Our framework reveals that agentic loops exhibit classifiable dynamics: \emph{contractive} (convergence toward stable semantic attractors), \emph{oscillatory} (cycling among attractors), or \emph{exploratory} (unbounded divergence).

Experiments on singular loops validate the framework.
Iterative paraphrasing produces contractive dynamics with measurable attractor formation and decreasing dispersion.
Iterative negation produces exploratory dynamics with no stable structure.
Crucially, \textbf{prompt design directly controls the dynamical regime}---the same model exhibits fundamentally different geometric behaviors depending solely on the transformation applied.

This work establishes that iterative LLM dynamics are \emph{predictable and controllable}, opening new directions for stability analysis, trajectory forecasting, and principled design of composite loops that balance convergence and exploration.
\end{abstract}

\section{Introduction}

Iterative LLM systems are now ubiquitous.
Self-Refine~\cite{madaan2023selfrefine} iteratively critiques and improves outputs.
Chain-of-thought~\cite{wei2022chain} decomposes reasoning into sequential steps.
Autonomous agents~\cite{yao2022react,shinn2023reflexion,park2023generative} maintain extended recursive interactions.
These systems share a common structure: \emph{agentic loops} where each output becomes the next input, $a_{t+1} = F(a_t)$.

Yet despite widespread deployment, a fundamental question remains unanswered: \textbf{what are the dynamics of these loops?}
Do they converge to stable outputs? Drift unboundedly? Oscillate between states?
Current work evaluates only \emph{endpoints}---task accuracy after $n$ iterations---while ignoring the \emph{trajectory} that produced them.
This gap is consequential: without understanding dynamics, we cannot predict when loops will stabilize, detect runaway divergence, or design loops with desired convergence properties.

\paragraph{The gap.}
Existing research treats iterations as independent steps rather than as a dynamical process.
Self-Refine~\cite{madaan2023selfrefine} measures task improvement but not geometric trajectory structure.
Agentic frameworks~\cite{yao2022react,shinn2023reflexion} demonstrate sophisticated behaviors without characterizing convergence or divergence patterns.
Embedding geometry research~\cite{ethayarajh2019contextual,wang2020understanding} analyzes static distributional properties but not how representations evolve under repeated transformations.

\paragraph{Contributions.}
We formalize agentic loops as discrete dynamical systems in semantic space, drawing on classical theory~\cite{strogatz2018nonlinear}.
Our contributions are:

\begin{enumerate}[label=(\roman*)]
    \item \textbf{Definitions.} Operational concepts of \emph{trajectory} (the sequence of semantic embeddings as the loop evolves), \emph{attractor} (stable regions detected via incremental clustering), and \emph{dynamical regime} (contractive, oscillatory, or exploratory). These are tailored to LLM transformations, not generic dynamical systems theory.

    \item \textbf{Measurement protocol.} A concrete pipeline from text to calibrated embedding to geometric indicators. We define and compute local drift (similarity to previous iteration), global drift (similarity to initial state), dispersion, and cluster persistence---all using calibrated similarity~\cite{tacheny2025calibrated} to ensure measurements align with human semantic judgments.

    \item \textbf{Control insight.} Empirical demonstration that \emph{prompt design determines the dynamical regime}. On the same LLM, iterative paraphrasing produces contractive dynamics (similarity increases, dispersion decreases, stable attractors form), while iterative negation produces exploratory dynamics (similarity fluctuates, no stable structure emerges).
    The transformation operator, not the model, controls the regime.
\end{enumerate}

\paragraph{Why it matters.}
These contributions enable capabilities absent from prior work:
\emph{stability analysis} (detect convergence/divergence from early iterations),
\emph{trajectory prediction} (forecast future semantic states),
and \emph{principled loop design} (engineer composite loops with controlled dynamics).
For computational creativity~\cite{boden2009computermodels,colton2012computational}, this provides infrastructure to balance exploration (novelty) and consolidation (value preservation).

\paragraph{Scope.}
For improved interpretability, we use a calibrated semantic similarity~\cite{tacheny2025calibrated} where measurements align with human judgments.
However, our definitions and qualitative results (contraction vs.\ divergence, attractor detection) are order-based and thus robust to any monotonic transformation of the similarity metric, including raw cosine similarity.

\paragraph{Claims investigated.}
\begin{itemize}
    \item C1: Singular loops induced by prompt P can fall into regimes measurable in the embedding space.
    \item C2: Under a paraphrase prompt, trajectories form persistent clusters with identifiable attractors and decreasing dispersion over time.
    \item C3: Under a negation prompt, no persistent cluster emerges under the tested similarity and dispersion thresholds.
\end{itemize}

\section{Related Work}
\label{sec:related}

\subsection{Iterative LLM Systems and Agentic Loops}

A growing body of work studies iterative and recursive LLM architectures.
Self-Refine~\cite{madaan2023selfrefine} demonstrated that LLMs can iteratively improve outputs through self-critique, measuring task accuracy gains across iterations.
Chain-of-thought prompting~\cite{wei2022chain} decomposes reasoning into sequential steps, evaluated by final answer correctness.
ReAct~\cite{yao2022react} interleaves reasoning with actions; Reflexion~\cite{shinn2023reflexion} uses verbal reinforcement for multi-trial improvement; generative agents~\cite{park2023generative} simulate human behavior through memory and reflection.

\textbf{What these works measure:} task performance, reward, success rate, output quality---all evaluated at the \emph{endpoint} of the iterative process.

\textbf{What they do not measure:} the \emph{trajectory} itself. How does semantic content evolve across iterations? Is the loop converging, oscillating, or drifting? At what rate? These questions are not addressed because no geometric measurement framework exists for this purpose.

\subsection{Dynamical Systems and Stability Analysis}

Discrete dynamical systems theory~\cite{strogatz2018nonlinear} provides mature tools for analyzing iterations $x_{t+1} = F(x_t)$: fixed points, periodic orbits, attractors, Lyapunov stability, contraction mappings.
These concepts are well-studied in control theory, optimization, and numerical analysis, where operators act on well-defined vector spaces with known analytic properties.

\textbf{Why this does not directly transfer:} LLM transformations differ fundamentally from classical settings.
The operator $F$ is a deep neural network with no closed-form expression.
The state space is not the embedding space itself but a discrete artifact space (text), which is only \emph{projected} into embeddings post-hoc for measurement.
The geometry is indirect, high-dimensional, and noisy.
Classical stability theorems assume conditions (Lipschitz bounds, differentiability) that cannot be verified for prompt-driven text transformations.

Consequently, while the \emph{vocabulary} of dynamical systems (trajectory, attractor, regime) is applicable, the \emph{measurement problem} is entirely new: how do we quantify these concepts when the operator is a black-box LLM and the state space is linguistic?

\subsection{Embedding Geometry}

Research on embedding geometry~\cite{ethayarajh2019contextual,wang2020understanding} has characterized \emph{static} properties of representation spaces: anisotropy, clustering, alignment-uniformity trade-offs.
This work analyzes the distribution of embeddings but not how representations \emph{evolve} under repeated transformations.
The temporal dimension---trajectories through embedding space---remains unexplored.

\subsection{Gap and Positioning}

The gap is now clear:
\begin{itemize}
    \item Iterative LLM systems exist and are widely deployed, but they are evaluated by \emph{outcomes}, not by the geometry of the transformation process.
    \item Dynamical systems theory provides the conceptual vocabulary, but its tools assume conditions that do not hold for LLM transformations.
    \item Embedding geometry characterizes static distributions, not temporal evolution.
\end{itemize}

\textbf{This work bridges these lines} by providing:
(i)~operational definitions of trajectory, attractor, and regime adapted to the LLM setting;
(ii)~a measurement protocol that maps text sequences to geometric indicators;
(iii)~empirical demonstration that these indicators reveal distinct, controllable dynamical behaviors.

Importantly, the notions of trajectory, attractor, and regime used here are not inherited from classical dynamical systems theory, but redefined operationally through observable, embedding-based indicators derived from discrete linguistic transformations.
The contribution is not the application of classical concepts to a new domain, but the construction of a \emph{measurement framework} that makes trajectory-level analysis possible where it previously was not.

\section{Semantic Space and Calibrated Similarity}
\label{sec:preliminaries}

This section describes the semantic measurement infrastructure used throughout this paper, following the calibration framework introduced in~\cite{tacheny2025calibrated}.

\paragraph{Artifact and representation spaces.}
We define the \textbf{artifact space} $\mathcal{A}$ as the set of all textual outputs that can be generated by a language model.
To enable geometric analysis, artifacts are projected into a continuous \textbf{representation space} $\mathcal{E} \subset \mathbb{S}^{d-1}$ via a normalized embedding function $\psi: \mathcal{A} \to \mathcal{E}$, where $\mathbb{S}^{d-1}$ is the unit hypersphere in $\mathbb{R}^d$.
Each artifact $a \in \mathcal{A}$ is represented by a unit vector $e = \psi(a) \in \mathcal{E}$.
The framework is embedding-model agnostic: any sentence encoder can instantiate $\psi$.

\paragraph{Calibrated similarity.}
Raw cosine similarity between embeddings suffers from \emph{anisotropy}: pretrained embedding spaces concentrate vectors in a narrow cone, causing similarities to cluster around high values even for semantically distant pairs~\cite{ethayarajh2019contextual}.
To address this, we employ the \textbf{calibrated similarity function} proposed in~\cite{tacheny2025calibrated}:
\begin{equation}
\tilde{s}(e_1, e_2) = f_{\mathrm{isotonic}}(\langle e_1, e_2 \rangle),
\end{equation}
where $f_{\mathrm{isotonic}}$ is a monotonic mapping learned via isotonic regression on human-annotated sentence pairs.
This calibration eliminates systematic bias, improves correlation with human judgments (Spearman $\rho \approx 0.86$) and preserves local stability under small linguistic perturbations ($\geq 98\%$).
Throughout this paper, all geometric measurements use the calibrated similarity $\tilde{s}$.

\paragraph{High-confidence similarity threshold.}
Following~\cite{tacheny2025calibrated}, we use the data-driven decision boundary $\tilde{\tau}_{\text{HCS}}$, the 5\% quantile of calibrated similarity among pairs with high human similarity.
This threshold guarantees that at least 95\% of truly similar pairs exceed this value and is used for cluster detection in subsequent sections.
Note that the specific value of $\tilde{\tau}_{\text{HCS}}$ depends on the embedding model; throughout this paper, we adopt the same embedding model as~\cite{tacheny2025calibrated}, yielding $\tilde{\tau}_{\text{HCS}} \approx 0.65$.

\paragraph{Robustness to metric choice.}
While we use calibrated similarity for improved interpretability, the qualitative conclusions of this work---contractive vs.\ exploratory dynamics, attractor formation, regime classification---depend only on \emph{order preservation}, not on the specific numeric values.
Since isotonic calibration is a monotonic transformation, it preserves pairwise similarity rankings and nearest-neighbor relationships.
Consequently, all definitions and empirical trends reported here would hold under raw cosine similarity or any other order-preserving metric.
The calibration improves human interpretability of threshold values but is not required for the framework's validity.

\section{Agentic Loop Dynamics: Theoretical Framework}
\label{sec:theoretical_framework}

This section formalizes the theoretical framework underlying agentic loop dynamics, defining key concepts such as trajectory, cluster and attractor before turning to empirical experiments.

\subsection{Primitive Operations}
\label{sec:primitive_operations}

Building on the artifact space $\mathcal{A}$, representation space $\mathcal{E}$, and calibrated similarity $\tilde{s}$ defined in Section~\ref{sec:preliminaries}, we introduce two fundamental operations:
\begin{itemize}
  \item $\mathrm{LLM} : \mathcal{A} \to \mathcal{A}$ denotes a large language model's generation operation, which takes a formatted input string (prompt) and produces a generated output string.
    The same underlying model may be invoked with different generation parameters (temperature $T$, top-p, top-k, seed, etc.), which we denote as $\mathrm{LLM}_{\theta}$ where $\theta$ represents the parameter configuration.
  \item $P : \mathcal{A} \to \mathcal{A}$ denotes a \emph{prompt template}, a function that embeds an artifact $a \in \mathcal{A}$ into a formatted prompt string suitable for LLM consumption.
    For example, $P(a) = \text{``Rewrite the following: \{$a$\}''}$ structures the artifact within an instruction.
\end{itemize}

\subsection{Agentic Loops and Trajectories}
\label{sec:definitions}

\paragraph{Agentic Loop.}
An \emph{agentic loop} is a discrete dynamical system $(\mathcal{A}, F)$ where $F : \mathcal{A} \to \mathcal{A}$ is a transformation operator that defines the recursive rule generating the sequence:
\begin{equation}
a_{t+1} = F(a_t), \quad t \in \mathbb{N}, \quad \text{with initial condition } a_0 \in \mathcal{A}.
\end{equation}
Using the primitive operations defined above, $F$ is implemented as:
\begin{equation}
F(a_t) = \mathrm{LLM}(P(a_t)).
\end{equation}
Each iteration produces a new artifact $a_{t+1}$ from the previous artifact $a_t$ through this composition of prompt template and LLM generation, starting from an initial artifact $a_0$.
The corresponding representations in the embedding space $\mathcal{E}$ evolve as:
\begin{equation}
e_t = \psi(a_t), \qquad e_{t+1} = \psi(a_{t+1}) = \psi(\mathrm{LLM}(P(a_t))), \quad \text{with } e_0 = \psi(a_0).
\end{equation}
The resulting sequence $\{e_t\}_{t=0}^T$ allows us to track the evolution of the loop dynamics in the continuous embedding space $\mathcal{E}$, where we can quantitatively measure distances and similarities between successive states.

We distinguish three types of agentic loops based on the structure of $F$:
\begin{enumerate}
  \item \textbf{Singular loop:} $F = \mathrm{LLM} \circ P$ where $P$ is a single prompt template.
    Each iteration applies one LLM call: $a_{t+1} = \mathrm{LLM}(P(a_t))$.
    This is the focus of our empirical study (Section~\ref{sec:experiments}).
  \item \textbf{Composite loop:} $F = \mathrm{LLM}_{n} \circ P_n \circ \cdots \circ \mathrm{LLM}_1 \circ P_1$ where each $(P_i, \mathrm{LLM}_i)$ pair represents a distinct transformation phase.
    Here, $\mathrm{LLM}_i$ denotes an LLM invocation with specific generation parameters (temperature, top-p, seed, etc.), which may differ across phases to induce different behaviors, for instance high temperature for creative exploration, low temperature for precise consolidation.
    Different model architectures can also be used (e.g., specialized models for critique vs. generation).
    For example, generate-critique-revise requires three sequential LLM calls with different prompt templates and potentially different sampling configurations. Composite loops enable controlled creativity through role-based phase alternation (Section~\ref{sec:creativity}).
  \item \textbf{Graph based loop:} $F$ is derived from a directed acyclic computation graph that could include parallel execution and conditional branches, forming the basis for multi-agent architectures (Section~\ref{sec:complex_architectures}).
\end{enumerate}
This paper establishes the theoretical framework applicable to all three types, then focuses empirically on \emph{singular loops} to rigorously characterize fundamental dynamical regimes.
Understanding singular loop behavior provides the necessary foundation for analyzing composite and graph based loops in future work.

\paragraph{Agentic Trajectory.}
The iterative process produces an ordered sequence of embedded representations:
\begin{equation}
\mathcal{T} = \{ e_t = \psi(a_t) \mid t = 0, 1, \dots, T \} \subset \mathcal{E},
\end{equation}
which we call the \emph{agentic trajectory}.
Each transition $(e_t, e_{t+1})$ represents a local semantic displacement, whose magnitude and direction quantify the instantaneous dynamics of the loop.

\subsection{Semantic Geometry of Trajectories}

Having defined how trajectories are generated, we now introduce geometric tools to measure their structure.
When a trajectory exhibits temporal stability, meaning embeddings remain close to each other for an extended period, we need mathematical concepts to characterize this clustering behavior.

\paragraph{Center of Gravity.}
Given a set of embeddings $E = \{e_{i_1}, e_{i_2}, \ldots, e_{i_k}\} \subset \mathcal{E}$, the \emph{center of gravity} $a_E$ is defined as the $L^2$-normalized mean:
\begin{equation}
a_E = \frac{\mu_E}{\|\mu_E\|_2}, \quad \text{where} \quad \mu_E = \frac{1}{k} \sum_{j=1}^{k} e_{i_j}.
\end{equation}
The center of gravity minimizes the sum of squared Euclidean distances to all embeddings in $E$ and is itself a unit vector: $\|a_E\|_2 = 1$.
This provides a natural reference point for measuring how tightly a set of embeddings is concentrated.

\paragraph{Semantic Dispersion.}
To quantify this concentration, we define the \emph{semantic dispersion}, which measures the maximum deviation from the center in calibrated similarity space:
\begin{equation}
\text{Dispersion}(E) = \max_{e \in E} \big[1 - \tilde{s}(e, a_E)\big],
\end{equation}
where $\tilde{s}$ is the calibrated similarity function from Section~\ref{sec:preliminaries}.
Smaller dispersion indicates tighter semantic coherence; $\text{Dispersion}(E) = 0$ means all embeddings are semantically identical to the center of gravity.
Together, the center of gravity and dispersion allow us to characterize when a trajectory remains in a stable semantic region.

\subsection{Clusters and Attractors}
\label{sec:clusters}

Armed with these geometric tools, we can now formalize the notion of a \emph{cluster}, a contiguous temporal segment where the trajectory maintains semantic stability.

\paragraph{Cluster.}
A \emph{cluster} $C \subseteq \mathcal{T}$ is a maximal contiguous temporal window $[t_a, t_b]$ of the agentic trajectory where embeddings maintain semantic coherence, up to a set of outliers $O_C$:
\begin{equation}
C = \{e_t \mid t \in [t_a, t_b] \setminus O_C\} \quad \text{with center of gravity} \quad a_C = \frac{\sum_{e \in C} e}{\|\sum_{e \in C} e\|_2}.
\end{equation}
The cluster represents a phase during which the loop operates within a bounded semantic region, repeatedly generating artifacts with similar meanings.

\paragraph{Cluster Validity Constraints.}
Not every temporal window qualifies as a cluster.
To ensure semantic coherence, a cluster must satisfy three validity constraints:
\begin{enumerate}
\item \textbf{Similarity ($\lambda$):} For all $t \in [t_a+1, t_b] \setminus O_C$, $\tilde{s}(e_{t-1}, e_t) \ge \lambda$, ensuring consecutive embeddings remain semantically close
\item \textbf{Dispersion ($\rho$):} $\text{Dispersion}(C) < \rho$, ensuring the cluster as a whole stays tightly concentrated
\item \textbf{Patience ($\kappa$):} At most $\kappa$ consecutive violations allowed before cluster termination, allowing temporary noise
\end{enumerate}
Parameters: $\lambda \in [0,1]$ (local coherence), $\rho > 0$ (global coherence), $\kappa \in \mathbb{N}$ (noise tolerance).
These constraints maintain semantic coherence while allowing for occasional outliers.

\paragraph{Attractor.}
For a cluster $C$ with center of gravity $a_C$, the \emph{attractor} is simply $a_C$, the semantic region toward which the trajectory converges during $C$'s temporal extent.
This terminology borrows from dynamical systems theory: the attractor represents the "pull" that keeps the trajectory within a stable semantic basin.

\subsection{Dynamic Regimes}
\label{sec:dynamic_regimes}

Finally, we characterize the global behavior of agentic loops by categorizing the qualitative patterns their trajectories exhibit over time.

\paragraph{Classification.}
An agentic loop's dynamics fall into one of three regimes:
\begin{itemize}
  \item \textbf{Contractive} ($\mathcal{R}_{\mathrm{ctr}}$): There exists a time $T$ such that for all $t \geq T$, the trajectory remains within a single cluster. The loop stabilizes around a coherent semantic region.
  \item \textbf{Oscillatory} ($\mathcal{R}_{\mathrm{osc}}$): The trajectory alternates among a finite set of recurrent clusters without converging to one. The loop cycles through distinct semantic regions.
  \item \textbf{Exploratory} ($\mathcal{R}_{\mathrm{exp}}$): The trajectory exhibits unbounded or aperiodic movement within $\mathcal{E}$, showing no stable attractor. The loop continuously generates novel semantic content without stabilization.
\end{itemize}
These categories provide a conceptual basis for describing the emergent behavior of agentic loops and for distinguishing structured evolution from uncontrolled semantic drift.

\paragraph{Scope and operationalization.}
The formal definitions provided above serve as conceptual anchors for the mathematical framework.
In our empirical analysis (Section~\ref{sec:experiments}), we use operational approximations rather than computing exact topological limits.
This pragmatic approach allows us to identify and characterize emergent structures within finite-sample trajectories while remaining grounded in the theoretical principles outlined above.

Note that the \emph{Oscillatory Regime} (periodic cycling among multiple attractors) is not observed in our experiments (Section~\ref{sec:experiments}).
The prompts we designed exhibit either full convergence (Contractive) or complete divergence (Exploratory), with no sustained oscillation between distinct semantic regions.
Exploring conditions that produce oscillatory dynamics, such as prompts with explicit alternating instructions or adversarial constraints, remains an avenue for future work.

\subsection{Assumptions and Scope}
\label{sec:assumptions}

The framework introduced in this work relies on a number of explicit assumptions
and scope restrictions, which we state here to clarify the validity and intended
interpretation of the proposed definitions and empirical observations.

\paragraph{A1. Black-box transformation operator.}
We assume that the transformation operator $F : \mathcal{A} \rightarrow \mathcal{A}$, implemented
via a prompt and a language model invocation, is treated as a black box.
No assumptions are made about its internal architecture, differentiability,
Lipschitz continuity, or contractive properties in the classical analytical sense.
All characterizations of dynamics are therefore observational rather than
analytical.

\paragraph{A2. Finite-horizon trajectories.}
Agentic trajectories are observed over a finite number of iterations $T$.
All notions of convergence, divergence, and stability are defined operationally
with respect to finite trajectories, rather than as asymptotic guarantees.
The framework does not claim formal convergence in the limit $T \rightarrow \infty$.

\paragraph{A3. Representation-based observation.}
While agentic loops operate in the discrete artifact space $\mathcal{A}$, their dynamics
are observed indirectly through a projection into a continuous representation
space $\mathcal{E}$ via an embedding function $\psi$.
We assume that relative semantic proximity in $\mathcal{E}$ reflects meaningful semantic
relationships between artifacts, but make no claim that $\mathcal{E}$ constitutes the true
state space of the generative process.

\paragraph{A4. Order-preserving similarity measures.}
The proposed definitions of clusters, attractors, and dynamical regimes rely on
relative similarity comparisons rather than absolute metric values.
Consequently, the framework assumes only that the similarity function preserves
the ordering of semantic proximity.
While calibrated similarity is used for interpretability, the qualitative
classification of regimes is invariant under any monotonic transformation of
the similarity measure.

\paragraph{A5. Prompt-defined operators and stochasticity.}
The effective transformation operator $F$ is defined jointly by the prompt
template and the language model decoding configuration.
Sampling stochasticity is therefore an inherent component of the dynamics.
Observed regimes characterize the statistical behavior induced by a given
prompt-model configuration, rather than deterministic properties of the model
alone.

These assumptions delineate the scope of the framework.
The objective is not to derive formal guarantees about language model dynamics,
but to provide an operational and interpretable methodology for characterizing
and comparing the observable behavior of iterative agentic loops.

\subsection{Cluster Detection Algorithm}
\label{sec:cluster_detection_algorithm}

To operationalize the cluster and attractor definitions presented in Section~\ref{sec:clusters}, we employ an \emph{incremental cluster detection algorithm} that processes the trajectory $\{e_t\}_{t=0}^{T}$ sequentially, identifying stable regions where embeddings maintain semantic coherence around a common center of gravity.
The algorithm implements the validity constraints defined earlier using three parameters: similarity threshold $\lambda \in [0,1]$, dispersion threshold $\rho > 0$ and patience $\kappa \in \mathbb{N}$.

\paragraph{Detection procedure.}
The algorithm processes embeddings sequentially and maintains a candidate cluster.
At each iteration $t$, it evaluates the cluster validity constraints:
\begin{enumerate}[label=(\roman*)]
  \item \textbf{Similarity:} consecutive embeddings satisfy $\tilde{s}(e_{t-1}, e_t) \ge \lambda$;
  \item \textbf{Dispersion:} adding $e_t$ preserves $\text{Dispersion}(C \cup \{e_t\}) < \rho$.
\end{enumerate}
If both constraints are satisfied, $e_t$ is added to the current cluster.
If either is violated, a \emph{violation counter} is incremented.
When the violation counter exceeds $\kappa$, the cluster is closed and its center of gravity is computed.
An iteration $t$ is classified as an \emph{outlier} if it occurs during an active cluster and violates at least one constraint, but consecutive violations have not yet exceeded the patience threshold $\kappa$.
Outliers are excluded from the cluster's center of gravity computation, but the cluster remains open until more than $\kappa$ consecutive violations occur.

This incremental procedure detects multiple clusters along a trajectory, each characterized by its center of gravity (attractor), temporal extent and semantic dispersion.

\subsection{Cluster Trajectory Visualization}
\label{sec:cluster_visualization}

To visualize cluster membership and semantic stability along an agentic trajectory, we construct a two-dimensional representation that encodes cluster identity, attractor similarity and outlier status simultaneously.

\paragraph{Graph construction.}
For a trajectory with $m$ detected clusters $\{C_1, \ldots, C_m\}$, each cluster $C_i$ is assigned a baseline at $y = i$.
For each iteration $t$ belonging to cluster $C_i$ (i.e., $t \in [t_a^i, t_b^i]$), we plot a point at coordinates:
\begin{equation}
(x, y) = \bigl(t,\, i + \alpha \cdot d(t, C_i)\bigr), \quad \text{where} \quad d(t, C_i) = 1 - \tilde{s}(e_t, a_i).
\end{equation}
Here, $a_i$ is the attractor of cluster $C_i$ and $\tilde{s}$ is the calibrated similarity.
The vertical offset $d(t, C_i)$ measures how far embedding $e_t$ deviates from the cluster's attractor in similarity space.
A scaling factor $\alpha > 0$ amplifies vertical displacements for better visualization of small deviations.

Points are colored by membership status:
\begin{itemize}
\item \textbf{Blue points:} valid cluster members ($t \in [t_a^i, t_b^i] \setminus O_i$), contributing to attractor computation
\item \textbf{Red points:} outliers ($t \in O_i$), excluded from attractor but within patience threshold $\kappa$
\end{itemize}

A grey band of height $\alpha \cdot \rho$ around each baseline $y = i$ visualizes the dispersion constraint: all blue points must remain below $y = i + \alpha \cdot \rho$ for the cluster to satisfy the dispersion constraint.

\paragraph{Reading the graph.}
The vertical position encodes semantic proximity to the attractor:
\begin{itemize}
\item A point at $y = i$ indicates $\tilde{s}(e_t, a_i) = 1$ (embedding identical to attractor)
\item A point at $y = i + \alpha \cdot 0.1$ indicates $\tilde{s}(e_t, a_i) = 0.9$ (high similarity)
\item Maximum vertical spread of blue points equals $\alpha \cdot \text{Dispersion}(C_i)$
\end{itemize}

Tight clusters exhibit blue points clustered near the baseline, while dispersed clusters show wider vertical spread.
Red outliers reveal iterations where the patience mechanism allows temporary violations without terminating the cluster.
Gaps between clusters indicate transition periods where the trajectory does not belong to any detected cluster.

\section{Experimental Study of Singular Agentic Loops}
\label{sec:experiments}

Having established the theoretical framework for analyzing agentic loop dynamics, we now turn to empirical validation.
As discussed in Section~\ref{sec:definitions}, singular loops represent the simplest agentic structure: a single transformation $F$ applied recursively.
By studying singular loops first, we can isolate fundamental dynamical phenomena (contraction, exploration, attractor formation) without the additional complexity introduced by multi-step compositions.
This foundational understanding is essential: composite loops, which alternate between generative and critical phases and graph based loops, which coordinate multiple interacting agents, both build upon the basic dynamics characterized here.
The experiments presented in this section demonstrate that even the simplest agentic loops exhibit rich, regime-dependent behavior, validating the geometric framework and establishing empirical benchmarks for future studies of more complex architectures.

\subsection{Experimental Design}

To empirically characterize the dynamics of agentic loops, we conduct a controlled experiment using two minimal \emph{singular} loops executed under identical computational conditions.  
Both operate in the artifact space $\mathcal{A}$ as purely textual transformations applied iteratively through a single LLM call per step.  
The objective is to observe whether distinct prompt formulations can induce fundamentally different dynamic regimes, specifically one exhibiting \emph{contractive} behavior and another exhibiting \emph{exploratory} divergence.

\paragraph{LLM configuration.}
All experiments are performed locally using the \texttt{Ollama} inference framework (version 0.12.9) with the \texttt{deepseek-r1:8b} model.
This model was selected for three reasons: (i)~it is small enough to run locally without GPU clusters, enabling full reproducibility; (ii)~despite its size, it demonstrates strong instruction-following capabilities on rewriting and reasoning tasks, ensuring that prompt-induced behaviors are faithfully executed; and (iii)~local execution eliminates API rate limits and guarantees deterministic infrastructure, which is essential when running 50-iteration loops repeatedly.
We use a sampling temperature of $0.8$, which introduces mild stochasticity while preserving semantic coherence, ensuring that the observed trajectories reflect both the model's internal tendencies and the intrinsic stability of the prompt formulation.
No explicit random seed was specified, so each generation uses one determined by the inference framework, contributing to the stochastic nature of the observed dynamics.

\paragraph{Motivation.}
The purpose of this experiment is not to model complex reasoning or task-solving behavior, but rather to reveal that even the simplest iterative linguistic transformations can produce structured dynamics when observed in the calibrated semantic space $(\mathcal{E}, \tilde{s})$.  
By isolating the loop mechanism to a single prompt applied recursively, we eliminate confounding factors such as multi-step planning or external memory accumulation.  
This enables direct observation of the intrinsic dynamical tendencies of the language model itself.

\paragraph{Experimental setup.}
Each loop is defined by a fixed prompt template $P$, into which the current artifact $a_t$ is inserted at each iteration:
\begin{equation}
a_{t+1} = F(a_t) = \mathrm{LLM}(P(a_t)).
\end{equation}
The process is repeated for $T = 50$ iterations, generating a trajectory $\{a_t\}_{t=0}^{T}$ and its projection $\{e_t = \psi(a_t)\}_{t=0}^{T}$ in the representation space $\mathcal{E}$.
This horizon balances computational cost against sufficient observation length: preliminary runs indicated that contractive loops stabilize within 20--30 iterations, while exploratory loops show no sign of convergence; $T=50$ thus provides adequate margin to distinguish these regimes.
All trajectories are embedded using the same function $\psi$ and analyzed in the calibrated similarity space defined in Section~\ref{sec:preliminaries}.

\paragraph{Initial condition.}
Both loops are initialized with the same starting sentence:
\begin{quote}
"Music has the power to connect people across cultures and generations."
\end{quote}
This neutral, semantically coherent sentence serves as the common baseline $a_0$ for comparing the contractive and exploratory dynamics.

\paragraph{Cluster detection configurations.}
To identify emergent structures such as clusters or attractors within each trajectory, we apply the incremental cluster detection algorithm described in Section~\ref{sec:cluster_detection_algorithm}.
We explore multiple parameter configurations $(\lambda, \rho, \kappa)$ to assess the robustness of attractor detection under varying spatial resolutions:
\begin{equation}
\begin{aligned}
(\lambda, \rho, \kappa) \in \{ &
(0.8, 0.1, 2),\,
(0.8, 0.2, 2),\,
(0.8, 0.3, 2)
\}.
\end{aligned}
\end{equation}
The similarity threshold $\lambda = 0.8$ enforces high local coherence, admitting only consecutive pairs that are clearly semantically related.
The dispersion radii $\rho \in \{0.1, 0.2, 0.3\}$ provide multi-resolution analysis: $\rho = 0.1$ detects only tightly packed micro-clusters, while $\rho = 0.3$ captures broader attractor basins; the increments are chosen to span interpretable fractions of the calibrated similarity scale.
The patience parameter $\kappa = 2$ tolerates brief noise (up to two consecutive violations) without prematurely terminating a cluster, while remaining strict enough to separate genuine regime transitions.
This grid allows both fine-grained and coarse-grained analyses of the clustering structure, ensuring that attractor detection is not an artifact of a single threshold choice.

\paragraph{Prompts and expected regimes.}
We define two archetypal prompts designed to induce contrasting dynamical behaviors in the same linguistic space.

\begin{itemize}
  \item \textbf{Contractive loop prompt:}
  \begin{quote}
  \small
  \texttt{You are a rewriting agent.  
  At each step, rewrite the sentence to make it sound slightly more natural and fluent,  
  while preserving the meaning exactly.  
  \\
  Current sentence: \{\{TEXT\}\}  
  \\
  Provide only the new sentence.}
  \end{quote}

  This formulation encourages small, directionally consistent improvements toward a stable expression, leading to decreasing semantic displacement over time.  
  We hypothesize that its trajectory in $\mathcal{E}$ will converge toward a unique attractor, a locally optimal point corresponding to the most fluent paraphrase of the initial sentence.

  \item \textbf{Exploratory loop prompt:}
  \begin{quote}
  \small
  \texttt{Summarize the current text in one sentence, then negate its main idea completely in an abstract way.  
  \\
  Current sentence: \{\{TEXT\}\}  
  \\
  Provide only the new sentence.}
  \end{quote}

  In contrast, this prompt enforces semantic inversion at each step, deliberately perturbing the meaning while maintaining grammatical coherence.  
  The expected effect is a divergent or oscillatory trajectory in $\mathcal{E}$, where successive iterations alternate between semantically distant regions without stabilization.
\end{itemize}

\paragraph{Analysis objectives.}
To characterize the emergent dynamics, we analyze each trajectory $\{e_t\}$ according to the following criteria:
\begin{enumerate}
  \item The sequence of calibrated similarities $\tilde{s}(e_t, e_{t-1})$, quantifying instantaneous dynamics.
  \item The evolution of the similarity $\tilde{s}(e_t, e_0)$ to the initial state, capturing long-term semantic drift or contraction.
  \item The detection and temporal persistence of \emph{clusters} and \emph{attractors} as defined in Section~\ref{sec:definitions}, revealing whether the loop settles into recurrent stable regions or exhibits perpetual divergence.
\end{enumerate}

A contractive loop is expected to produce a single dense cluster of high internal similarity, corresponding to a stable attractor.  
Conversely, an exploratory loop should either fail to produce persistent clusters or exhibit transient, disjoint attractors, reflecting alternating semantic polarity and non-convergent trajectories.
These contrasting behaviors form the empirical foundation for the study of dynamics and stability in subsequent analyses.

\subsection{Contractive Loop Dynamics}

This subsection presents the empirical results obtained from the contractive loop.
For each configuration of the clustering parameters $(\lambda, \rho, \kappa)$, we analyze the resulting trajectory in the calibrated embedding space $(\mathcal{E}, \tilde{s})$, focusing on the emergence of clusters, attractors and inter-cluster similarity patterns.

The contractive loop was designed to iteratively refine a sentence while preserving its semantic meaning.
As expected, the observed trajectory rapidly converges toward a small, dense region in $\mathcal{E}$, revealing a clear attractor structure.
We first analyze the overall geometric behavior of the trajectory, independent of any clustering configuration, before examining how different density thresholds $(\lambda, \rho, \kappa)$ reveal the underlying attractor organization.
The complete sequence of generated artifacts is provided in Appendix~\ref{app:contractive_trajectory} for reference.

\subsubsection{Global and Local Dynamics}

\begin{figure}[h]
  \centering
  \includegraphics[width=\linewidth]{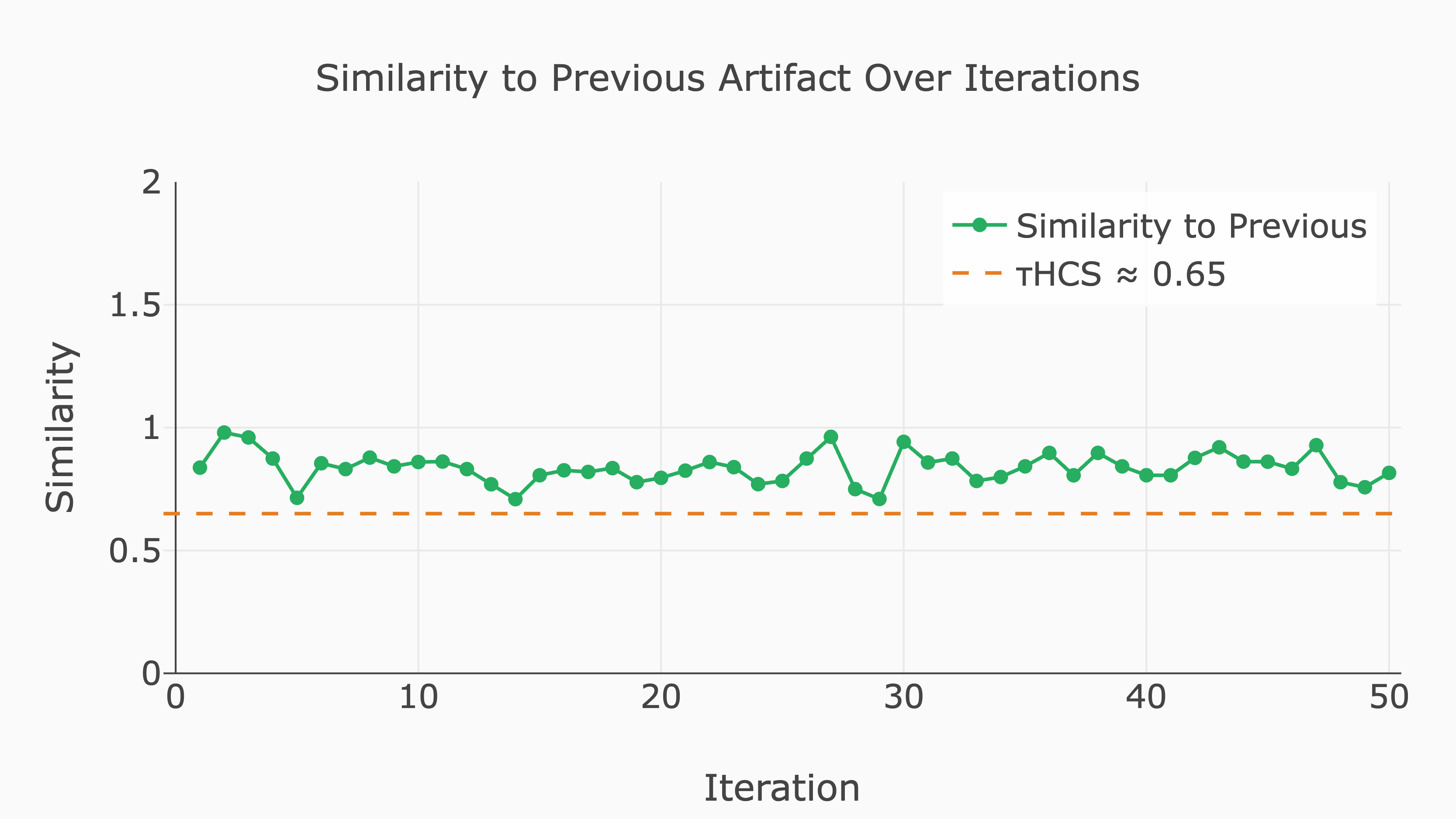}
  \caption{Local similarity $\tilde{s}(e_t, e_{t-1})$ for the contractive loop. Values remain consistently high ($>0.8$), indicating small semantic displacements between consecutive iterations.}
  \label{fig:contractive_local}
\end{figure}

\paragraph{Local stability.}
The calibrated similarity $\tilde{s}(e_t, e_{t-1})$ between consecutive embeddings remains consistently high, in the range $0.82$–$0.95$ (Figure~\ref{fig:contractive_local}).
Each rewriting step introduces only minimal semantic changes.

\begin{figure}[h]
  \centering
  \includegraphics[width=\linewidth]{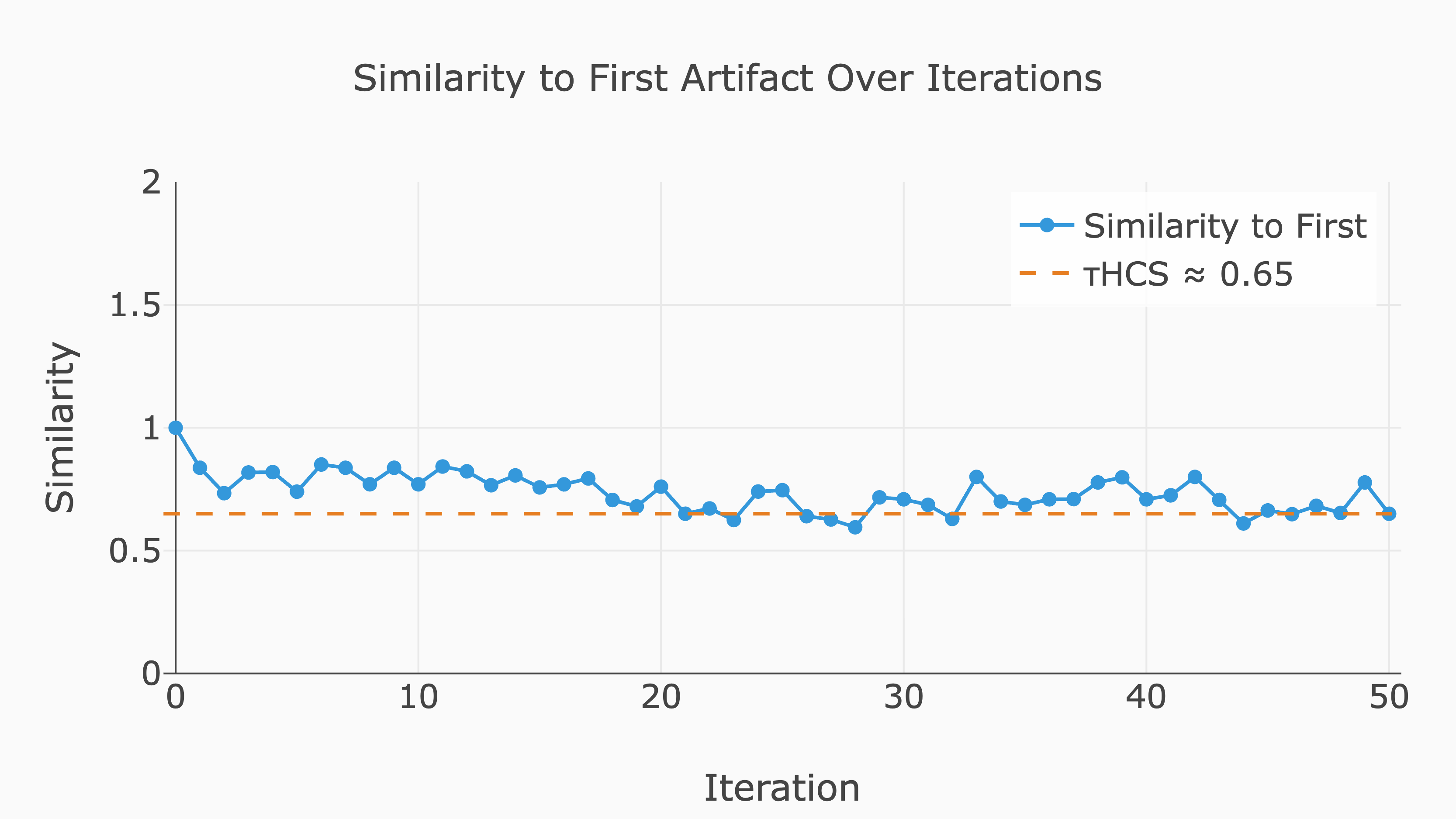}
  \caption{Global similarity $\tilde{s}(e_t, e_0)$ for the contractive loop. The trajectory stabilizes at moderate similarity to the initial state, indicating bounded drift.}
  \label{fig:contractive_global}
\end{figure}

\paragraph{Global convergence.}
The similarity to the initial embedding $\tilde{s}(e_t, e_0)$ stabilizes near $0.75$ after initial iterations (Figure~\ref{fig:contractive_global}).
The trajectory stops drifting and remains confined within a bounded region of $\mathcal{E}$.

\paragraph{Interpretation.}
The consistency between local and global patterns shows that the operator $F$ exhibits local contractivity in practice, despite no formal guarantee on $\psi$'s continuity.
The contractive loop thus defines a smooth trajectory approaching an equilibrium region, setting the stage for a finer-grained analysis of its attractor structure through clustering.
These observations are consistent with $\mathcal{R}_{\mathrm{ctr}}$ as formally defined in Section~\ref{sec:dynamic_regimes}: the trajectory exhibits behavior consistent with convergence toward a stable region in $\mathcal{E}$, approximating a fixed point attractor.

\subsubsection{Cluster and Attractor Analysis}

We now apply the density-based clustering framework defined in Section~\ref{sec:definitions} to characterize how the trajectory organizes into stable subregions or attractors.
Three clustering configurations were tested, varying only the spatial radius $\rho \in \{0.1, 0.2, 0.3\}$ while keeping $\lambda = 0.8$ and $\kappa = 2$ constant.
We present the analysis for $\rho = 0.2$, which provides an intermediate resolution that clearly reveals the attractor structure; complete results for all configurations are provided in Appendix~\ref{app:robustness_rho}.

\paragraph{Cluster structure at $\rho = 0.2$.}

At this resolution, the clustering algorithm identifies two persistent clusters corresponding to distinct stabilization phases: an initial transient phase (iterations 0--18) and a final convergence basin (iterations 21--50).

\begin{table}[h]
\centering
\begin{tabular}{lccc}
\toprule
\textbf{Cluster} & \textbf{Iteration Range} & \textbf{Dispersion} & \textbf{Outliers} \\
\midrule
Cluster 1 & 0–18  & 0.1773 & 6 \\
Cluster 2 & 21–50 & 0.1500 & 9 \\
\bottomrule
\end{tabular}
\vspace{0.5em}
\caption{Cluster configuration for the contractive loop under $(\lambda, \rho, \kappa) = (0.8, 0.2, 2)$.}
\label{tab:contractive_config_0.2}
\end{table}

\begin{figure}[h]
  \centering
  \includegraphics[width=\linewidth]{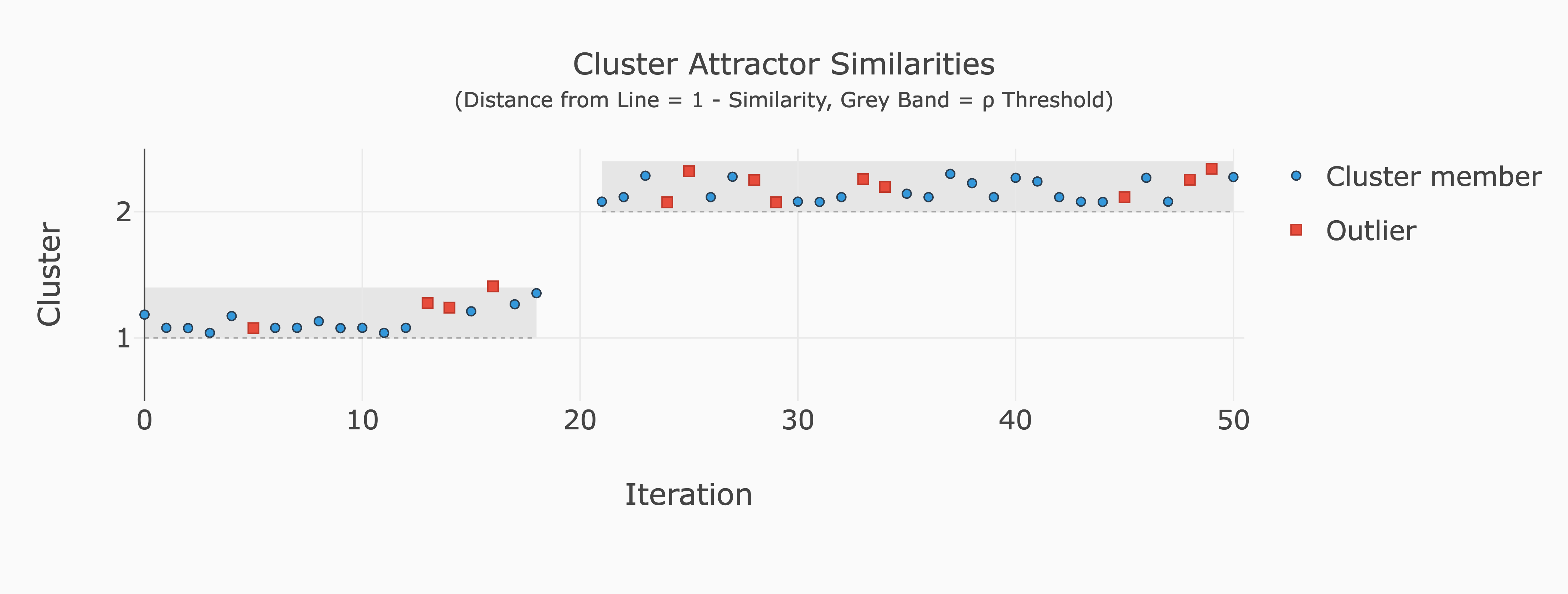}
  \caption{Cluster membership timeline for the contractive loop under $(\lambda, \rho, \kappa) = (0.8, 0.2, 2)$.
  Each point represents an iteration, color-coded by cluster membership.
  Outliers (in red) correspond to transient points at cluster boundaries.}
  \label{fig:contractive_cluster_timeline_0.2}
\end{figure}

The temporal organization of the trajectory, shown in Figure~\ref{fig:contractive_cluster_timeline_0.2}, confirms a single transition between two stable basins around iteration~20.
The early iterations (Cluster~1) exhibit slightly higher dispersion reflecting initial semantic exploration, while the second phase (Cluster~2) displays long persistence and decreasing internal dispersion---hallmarks of convergence toward a stable attractor.

\paragraph{Multi-resolution robustness.}
Identical qualitative behavior is observed across all tested resolutions: at $\rho = 0.1$ (finer resolution), five micro-clusters emerge that merge smoothly over time; at $\rho = 0.3$ (coarser resolution), a single cluster encompasses the entire trajectory.
In all cases, clusters exhibit temporal persistence, high inter-cluster similarity ($>0.78$), and monotonically decreasing dispersion.
See Appendix~\ref{app:robustness_rho} for complete tables, figures, and similarity matrices.

\paragraph{Summary.}
The cluster analysis confirms that the contractive loop satisfies the empirical criteria for $\mathcal{R}_{\mathrm{ctr}}$ (Section~\ref{sec:dynamic_regimes}):
\begin{itemize}
  \item persistent clusters with identifiable attractors;
  \item decreasing dispersion indicating convergence;
  \item no oscillation or divergence across 50 iterations.
\end{itemize}
The consistency of these findings across three clustering resolutions---without parameter tuning---strengthens the conclusion that the observed convergence reflects genuine geometric properties of the trajectory rather than artifacts of a particular configuration.
However, these observations are based on $T=50$ iterations; while the trajectory shows no signs of leaving the attractor within this window, proving long-term stability rigorously would require analyzing the probability of entering and remaining within a cluster as a function of the transformation operator $F$ and the embedding geometry.

\subsection{Exploratory Loop Dynamics}

In contrast to the previous rewriting loop, the exploratory loop was designed to alternate between summarization and semantic negation.
This alternation introduces an intrinsic instability in the semantic space, continuously disrupting coherence and preventing any convergence toward a stable region of $\mathcal{E}$.
The resulting trajectory exhibits high variability and recurrent semantic inversions, typical of a divergent or oscillatory dynamic.
The complete sequence of generated artifacts is provided in Appendix~\ref{app:exploratory_trajectory} for reference.

\subsubsection{Global and Local Dynamics}

\begin{figure}[h]
  \centering
  \includegraphics[width=\linewidth]{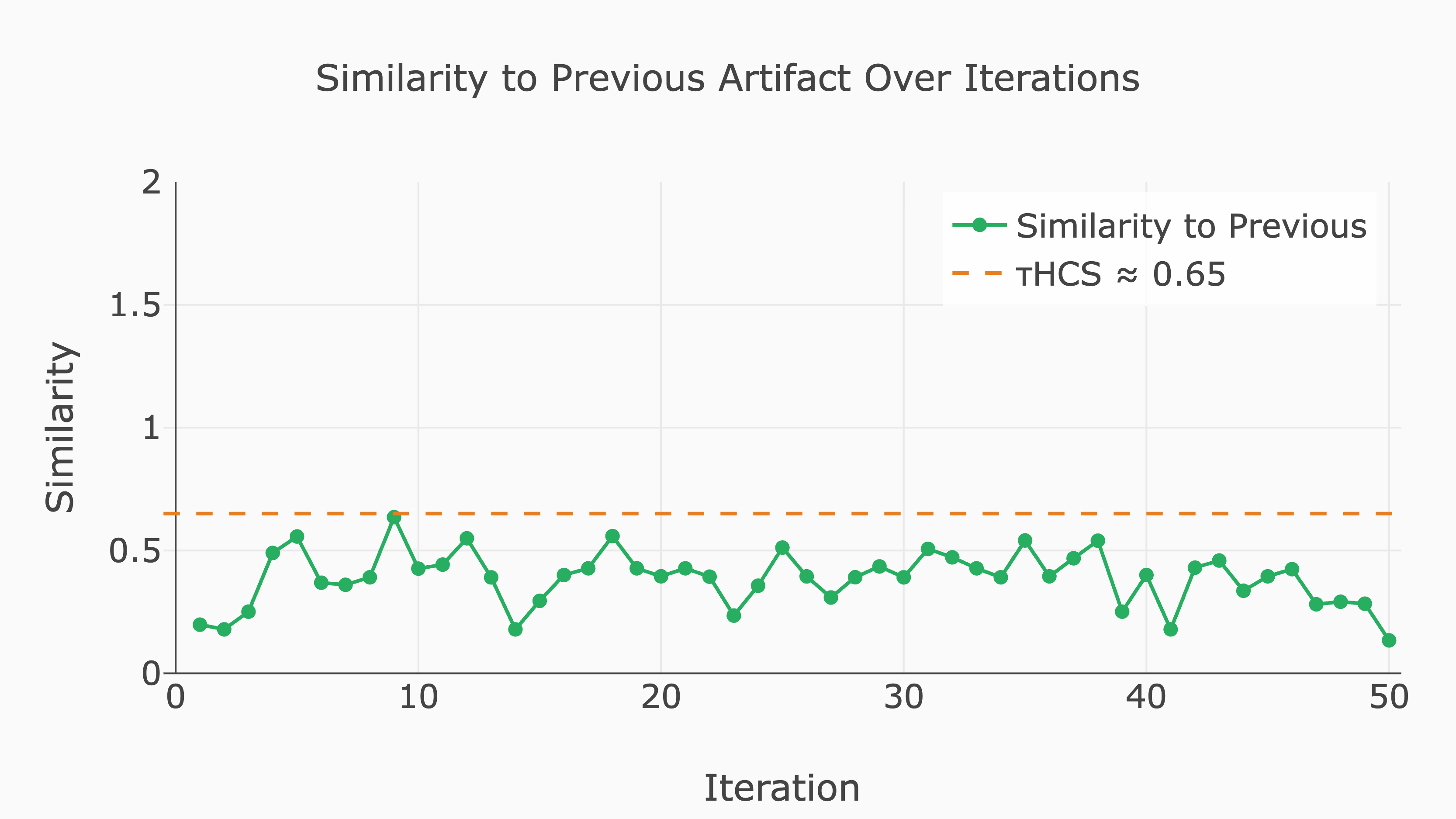}
  \caption{Local similarity $\tilde{s}(e_t, e_{t-1})$ for the exploratory loop. Values fluctuate erratically, indicating large semantic displacements at each iteration.}
  \label{fig:exploratory_local}
\end{figure}

\paragraph{Local instability.}
The local similarity $\tilde{s}(e_t, e_{t-1})$ fluctuates widely between $0.2$ and $0.6$, occasionally dropping close to zero (Figure~\ref{fig:exploratory_local}).
Each step produces a major semantic reorientation rather than incremental refinement.

\begin{figure}[h]
  \centering
  \includegraphics[width=\linewidth]{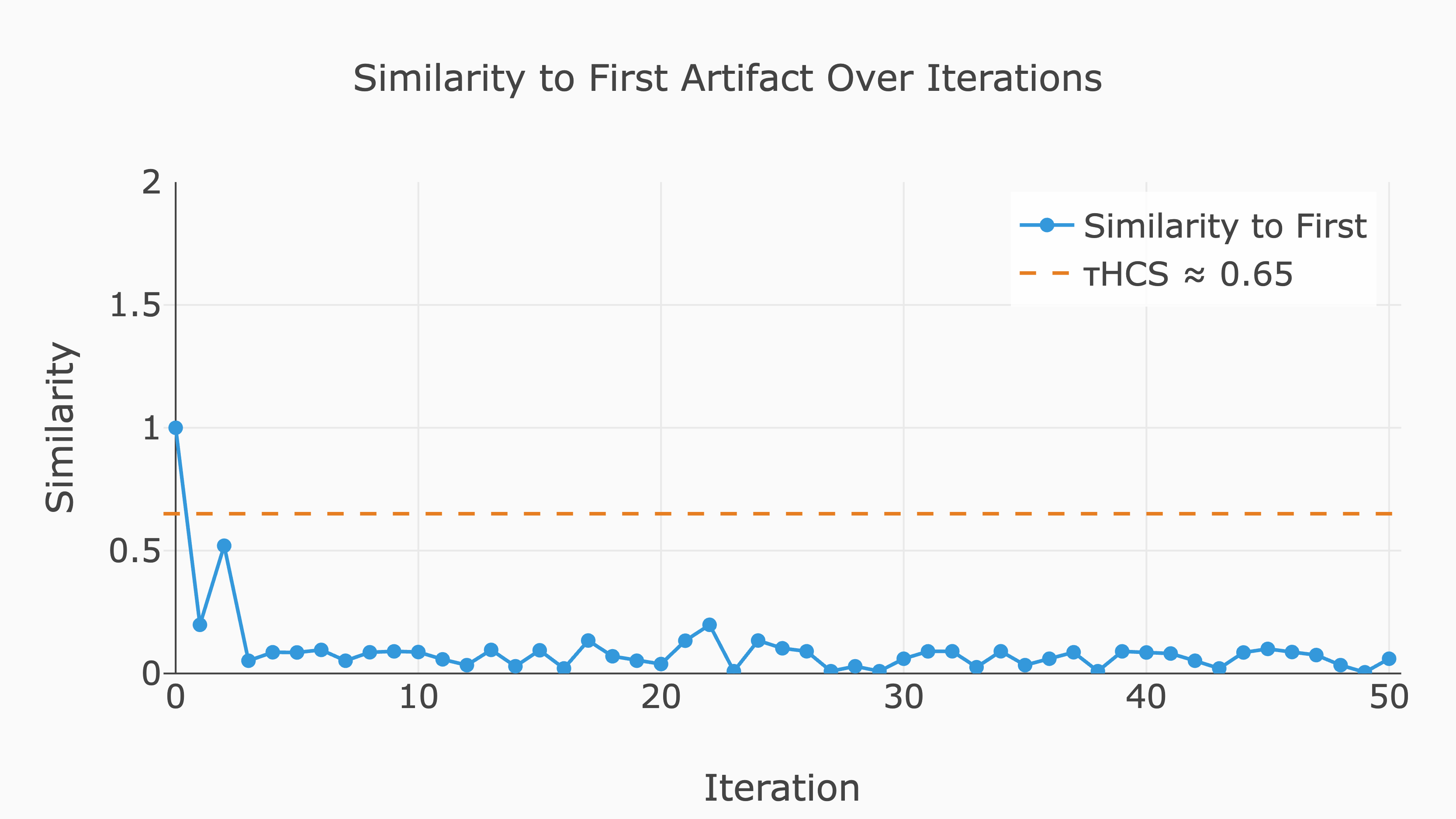}
  \caption{Global similarity $\tilde{s}(e_t, e_0)$ for the exploratory loop. The trajectory rapidly diverges from the initial state and remains semantically distant throughout.}
  \label{fig:exploratory_global}
\end{figure}

\paragraph{Global divergence.}
The similarity to the initial embedding $\tilde{s}(e_t, e_0)$ remains very low throughout, mostly below $0.2$ (Figure~\ref{fig:exploratory_global}), indicating continuous generation of artifacts semantically unrelated to the starting point.

\paragraph{Interpretation.}
The exploratory loop behaves as a \emph{divergent operator} in $(\mathcal{E}, \tilde{s})$.
Each iteration projects the embedding into a distant and nearly orthogonal region of the space, breaking any potential contractive pattern.
The trajectory thus explores disconnected semantic areas rather than converging toward an equilibrium.
This persistent divergence and absence of convergence align with $\mathcal{R}_{\mathrm{exp}}$ as formally defined in Section~\ref{sec:dynamic_regimes}: unbounded or aperiodic movement within $\mathcal{E}$ with no stable attractor.
As with the contractive case, these observations are limited to $T=50$ iterations. We cannot formally exclude the possibility that a cluster might eventually form over longer horizons. However, the consistently high inter-iteration distances suggest this is unlikely. A rigorous characterization would require a mathematical framework for analyzing the probability of cluster formation under a given transformation operator.

\subsubsection{Clustering Attempt and Absence of Attractors}

To test whether transient structures could still emerge locally, we applied the same density-based clustering procedure as for the contractive loop.
However, no valid cluster was detected under any tested configuration.
This complete absence of clusters directly confirms $\mathcal{R}_{\mathrm{exp}}$ behavior (Section~\ref{sec:dynamic_regimes}): the trajectory exhibits no stable attractor basin.

This failure to identify clusters reflects the fundamental instability of the exploratory process:
the embeddings never remain within a stable neighborhood long enough to satisfy the density or similarity conditions required for cluster formation.

\paragraph{Empirical evidence.}
Throughout the 50 iterations, each embedding violates the spatial proximity criterion $\rho$ almost immediately after creation.  
The lack of recurrence, combined with high inter-step distances and very low similarities, results in a completely dispersed point cloud in $\mathcal{E}$.

\paragraph{Geometric interpretation.}
The absence of attractors implies that the transformation function $F$ induced by this loop does not define a bounded trajectory.  
Instead, $F$ exhibits a form of \emph{semantic exploration}: each output resets the contextual reference frame, pushing the system toward a new region of the manifold that bears little relation to the previous one.

\paragraph{Summary.}
The exploratory loop displays persistently low local similarities ($\tilde{s}(e_t, e_{t-1}) < 0.5$), no geometric stabilization either locally or globally, and complete absence of clusters or attractor basins for all tested parameters.
In semantic terms, this loop corresponds to continuous destruction and regeneration of meaning: each iteration rewrites the narrative space from scratch.

\subsection{Comparative Summary}

The empirical analysis of contractive and exploratory loops reveals fundamentally distinct dynamical regimes, following the formal definitions established in Section~\ref{sec:dynamic_regimes}.
Table~\ref{tab:loop_comparison} summarizes the key differences.

\begin{table}[htbp]
\centering
\small
\begin{tabular}{lcc}
\toprule
\textbf{Metric} & \textbf{Contractive Loop} & \textbf{Exploratory Loop} \\
\midrule
Mean local similarity $\tilde{s}(e_t, e_{t-1})$  & $>$0.85          & $<$0.5           \\
Global similarity $\tilde{s}(e_t, e_0)$          & $\sim$0.75       & $<$0.2           \\
Clusters detected             & 5 $\to$ 1        & 0                \\
Regime type                   & Contractive       & Exploratory        \\
\bottomrule
\end{tabular}
\vspace{0.5em}
\caption{Comparative summary of contractive and exploratory agentic loop dynamics.}
\label{tab:loop_comparison}
\end{table}

These results provide empirical evidence for the existence of distinct dynamic regimes in recursive generative systems.
The contractive loop exhibits progressive semantic refinement, converging toward a stable attractor as evidenced by high local similarity ($>0.85$) and bounded global drift.
In contrast, the exploratory loop exhibits persistent instability, with each iteration producing semantically distant outputs ($\sim$1.0 Euclidean distance) and no detectable cluster structure.
These findings substantiate Claims C1–C3.
This dichotomy confirms that the choice of prompt and transformation operator fundamentally determines the long-term behavior of agentic trajectories, ranging from stable convergence to unbounded semantic wandering.

\section{Discussion and Perspectives}
\label{sec:discussion}

The empirical results presented above highlight two extreme regimes of agentic dynamics within the embedding manifold $(\mathcal{E}, \tilde{s})$:  
$\mathcal{R}_{\mathrm{ctr}}$, characterized by convergence toward a stable attractor, and $\mathcal{R}_{\mathrm{exp}}$, marked by unbounded divergence and the absence of any stable region.  
These two archetypal behaviors constitute the endpoints of a broader spectrum of agentic dynamics that could potentially be modulated, combined or even harnessed for creative control.

\subsection{Limitations}

The present study has several limitations that scope its conclusions:
\begin{itemize}
    \item All experiments use a single LLM (\texttt{deepseek-r1:8b}) with fixed decoding settings (temperature $0.8$); different architectures or sampling strategies may yield different dynamical signatures.
    \item Trajectories are observed through a single embedding function $\psi$ and calibrated similarity $\tilde{s}$; alternative embeddings or metrics could reveal different geometric structures.
    \item The observation horizon is limited to $T=50$ iterations, which may not capture slower convergence phenomena or late-emerging instabilities.
    \item The two prompts were chosen as archetypal extremes (pure paraphrase, pure negation) to cleanly separate regimes; they do not constitute an exhaustive map of the space of possible dynamical behaviors.
\end{itemize}

\subsection{Model Dependence and Generalization}

A first natural extension concerns the robustness of these observations across different embedding models and language models.

\paragraph{Embedding representations.}
The current study relies on a single embedding function $\psi$ to project textual artifacts into $\mathcal{E}$.  
However, various embedding architectures exhibit distinct geometric anisotropies and similarity landscapes.  
Future work should therefore replicate the same agentic trajectories across multiple embedding backbones to evaluate whether contraction and exploration are intrinsic to the generative process $F$ or contingent upon the metric structure of $\mathcal{E}$.

\paragraph{Language models.}
Similarly, the generation operator $F$ depends on the underlying LLM and its internal sampling mechanisms.  
Different architectures (e.g., GPT, Claude, Mistral, Llama) and decoding settings (temperature, top-$p$, beam search) may alter dynamics of $F$.  
A comparative study could determine whether exploratory dynamics arise from semantic inversion per se or from model-specific sampling noise.

\subsection{Toward Composite Agentic Loops}

Beyond the study of isolated transformations, a promising avenue is the investigation of \emph{composite loops}, where contractive and exploratory iterations alternate within a single process.  
Such compositions could emulate a controlled exploration–consolidation cycle:
\begin{equation}
F_{\text{composite}} = F_{\text{contractive}} \circ F_{\text{exploratory}} \quad \text{or} \quad F_{\text{exploratory}} \circ F_{\text{contractive}}.
\end{equation}
Intuitively, an exploratory iteration could inject novelty and creative divergence, while a subsequent contractive step would restore semantic coherence and bring the system back toward a stable region.  
This dual-phase regime might enable the construction of agents that balance \emph{creative drift} and \emph{semantic control}, producing bounded yet innovative trajectories within $\mathcal{E}$.

The quantitative analysis of such mixed loops would require monitoring:
\begin{itemize}
  \item the alternation between expansion (entropy increase) and contraction (entropy reduction);
  \item the long-term recurrence or periodicity of the resulting trajectory;
  \item and the net displacement relative to the initial semantic manifold.
\end{itemize}

\subsection{Toward Geometry of Agentic Cognition}
\label{sec:cognition}

The trajectory analysis framework developed in this paper enables a natural question: can an agentic loop \emph{reason about its own geometric dynamics} and adapt its behavior accordingly?
Currently, trajectory control remains external---modulated through prompt engineering or sampling parameters (temperature, top-p) specified a priori by a human engineer.

A fundamentally different paradigm, which we term \textbf{Geometry of Agentic Cognition}, would enable the loop itself to observe its trajectory geometry (local similarity, global drift, dispersion), reason about whether its current regime aligns with the objective (convergence toward a solution vs.\ exploration of alternatives), and adapt its strategy accordingly.
This would require trajectory introspection (embedding recent outputs and computing geometric indicators within the loop context) and geometric prompt characterization (pre-classifying prompts by their dynamical properties to enable informed selection).
Such systems would transform from passive trajectory executors into \emph{geometric navigators} that actively steer through semantic space.
The framework established here provides the measurement infrastructure for investigating whether agentic systems can achieve such geometric self-regulation.

\subsection{Speculative Application: Creativity}
\label{sec:creativity}

The geometric indicators developed here---drift, dispersion, attractor formation---bear a suggestive correspondence to dimensions of computational creativity as formalized by Boden~\cite{boden2009computermodels} and Colton \& Wiggins~\cite{colton2012computational}.
Exploratory dynamics (high dispersion, no stable attractors) may relate to \emph{novelty generation}, while contractive dynamics (convergence, low dispersion) may relate to \emph{value consolidation}.

This suggests a speculative direction: composite loops that alternate exploratory and contractive phases could potentially balance novelty and coherence, with prompt design and temperature modulating the regime at each phase~\cite{zhao2024creativity}.
We emphasize that this remains a hypothesis for future investigation; the present work provides only the measurement infrastructure, not empirical validation of creativity-specific claims.

\subsection{Extending Toward Complex Agentic Architectures}
\label{sec:complex_architectures}

Finally, the framework introduced here can be generalized beyond singular loops to more complex agentic architectures composed of interacting subloops, each operating under distinct dynamics and objectives.  
Examples include:
\begin{itemize}
  \item hierarchical loops, where upper layers modulate the contractive or exploratory tendency of lower layers;
  \item cooperative loops, in which multiple agents operate in parallel within shared embedding subspaces;
  \item and adversarial or self-reflective loops, where one agent destabilizes while another re-stabilizes the shared representation.
\end{itemize}
Such architectures could reveal emergent phenomena such as collective attractors, dynamic equilibrium points or oscillatory creative states, bridging the gap between low-level generative processes and higher-order cognitive modeling.

\paragraph{Outlook and Future Directions.}
By formalizing agentic dynamics in terms of their geometric behavior within embedding spaces, this work opens the path to a quantitative study of creativity and stability in language models.
Future research on composite, parameterized and hierarchical agentic loops may ultimately lead to the design of \emph{controllable creative agents}: systems capable of exploring without drifting and stabilizing without stagnating.
This discussion opens the way to broader formal and experimental investigations, which are summarized in the final conclusion.

\section{Conclusion}
\label{sec:conclusion}

This work introduced a geometric framework for analyzing the dynamics of agentic loops: recursive processes in which large language models iteratively transform their own outputs.
By grounding these dynamics in a calibrated semantic embedding space $(\mathcal{E}, \tilde{s})$, we made it possible to quantify how iterative LLM processes unfold, stabilize or diverge.

\paragraph{Summary of contributions.}
\begin{enumerate}
  \item A formal theoretical framework distinguishing artifact space $(\mathcal{A}, F)$ from representation space $(\mathcal{E}, \|\cdot\|)$, with precise definitions of trajectories, clusters, attractors and dynamical regimes.
  \item An incremental cluster detection algorithm that operationalizes these concepts for empirical analysis.
  \item Empirical demonstration that agentic loops exhibit distinct dynamical regimes (contractive or exploratory) depending on prompt formulation, with measurable geometric signatures.
\end{enumerate}

\paragraph{Empirical findings.}
\begin{itemize}
  \item The \textbf{contractive loop} produces a bounded trajectory converging toward a dense attractor region, with local contractivity and persistent clustering.
  \item The \textbf{exploratory loop} generates an unbounded trajectory with no persistent clusters, characterized by high local displacements and continuous semantic drift.
\end{itemize}
These regimes define a spectrum from convergent stabilization to divergent exploration.

\paragraph{Theoretical implications.}
The observations are consistent with a view of LLM transformations as mappings whose dynamics resemble those governed by an effective contraction rate.
While such bounds cannot be measured directly for deep generative operators, observed trajectory patterns suggest a regime-dependent effective contraction rate that could be estimated through embedding-space analysis.

\paragraph{Broader outlook.}
The contrast between contractive and exploratory regimes highlights a key tension: the need to balance \emph{semantic exploration} and \emph{stability}.
Composite loops that alternate exploratory and contractive phases offer a systematic path toward \emph{controlled agentic behavior}: exploring semantic space while maintaining coherence.
This connects naturally to computational creativity research, where the geometric indicators developed here can operationalize theoretical notions of novelty, value and surprise.

More broadly, this work establishes a bridge between agentic AI and dynamical systems theory, encouraging future study of how autonomous generative processes evolve within semantic embedding spaces.

\section*{Acknowledgments}
The author thanks open-source contributors of the \texttt{Ollama} and \texttt{Deepseek} projects for enabling reproducible experimentation on local models, the \texttt{Xenova} project for providing a reproducible transformer inference pipeline and the creators of the MTEB STS Benchmark dataset for providing calibrated human similarity judgments.

\section*{Statements and Declarations}

\subsection*{Competing Interests}
The author declares that there are no competing financial or non-financial interests that could have influenced the work reported in this paper.

\subsection*{Data Availability}
The experiments were conducted using custom implementations. Code availability will be considered following the peer-review process.

\bibliographystyle{unsrt}
\bibliography{references}

\appendix

\section{Contractive Loop: Complete Trajectory}
\label{app:contractive_trajectory}

This appendix provides the complete sequence of artifacts generated by the contractive loop experiment.
The initial artifact $a_0$ undergoes 50 iterations of the rewriting transformation.

The progressive evolution demonstrates contractive dynamics: initial rapid refinement (iterations 0--20), followed by stabilization into a dense semantic attractor (iterations 20--50) where only minor stylistic variations occur while the core message remains fixed.

\begin{small}
\begin{verbatim}
t=0:  Music has the power to connect people across cultures and
      generations.

t=1:  Music has the remarkable ability to bridge cultural divides
      and unite people across different eras.

t=2:  Music has a remarkable ability to bridge cultural divides and
      unite people across different eras.

t=3:  Music has an extraordinary power to bridge cultural divides
      and bring people together across different eras.

t=4:  Music possesses a remarkable ability to transcend cultural
      boundaries and unite individuals across various time periods.

t=5:  Music has a remarkable power to bridge cultural divides and
      unite people from different time periods.

t=6:  Music has an extraordinary ability to transcend cultural
      boundaries and bring together individuals from diverse
      backgrounds, past and present.

t=7:  Music's unique power allows it to bridge cultural divides and
      unite people from varied backgrounds, both historically and
      currently.

t=8:  Music has a profound ability to transcend cultural boundaries,
      bringing together individuals from diverse backgrounds, past
      and present.

t=9:  Music has a remarkable power to bridge cultural divides,
      connecting people across time and geography.

t=10: Music has an extraordinary ability to transcend cultural
      boundaries, uniting people across space and history.

t=11: Music possesses an extraordinary power to bridge cultural
      divides, unifying people across time and geography.

t=12: Music has the remarkable ability to transcend cultural
      boundaries, uniting people across different times and places.

t=13: Music's incredible capacity to bridge cultural divides has
      unified people throughout history and across the globe.

t=14: Music has long been able to bring people together, uniting
      cultures from all over the world.

t=15: Music has brought people together for centuries, bridging
      cultural divides across the globe.

t=16: For centuries, music has been a powerful force in bringing
      people together, regardless of cultural background or
      geographical location.

t=17: For centuries, music has played a unifying role across cultures
      and distances, fostering connections among people worldwide.

t=18: Music has been a unifying force for centuries, bridging
      cultural divides and connecting people globally through its
      universal language.

t=19: Music has long served as a powerful unifier, transcending
      cultural boundaries to unite people across the globe through
      its universally understood language.

t=20: Music has been a potent force for unity, bridging cultural
      divides and connecting people worldwide through a shared
      emotional vocabulary.

t=21: Music has long been a powerful unifying force that transcends
      cultural boundaries, speaking to our shared human emotions in
      a universal language.

t=22: Music has been a unifying force for centuries, capable of
      bridging cultural divides and speaking directly to our common
      humanity through its universal language.

t=23: Music has been a powerful unifying force throughout history,
      transcending cultural boundaries and speaking directly to our
      shared human experience through its universal language.

t=24: Throughout history, music has served as a profoundly unifying
      force that has bridged cultural divides and connected people
      on a deep, universal level, tapping into the common human
      emotions it evokes.

t=25: Throughout history, music has played a profoundly unifying
      role, uniting cultures and connecting people across profound
      emotional lines that tap into the deepest human experiences.

t=26: Music has been a profoundly unifying force throughout history,
      bridging cultural divides and tapping into universal emotions
      that resonate with our very core.

t=27: Throughout history, music has played a profoundly unifying
      role, bridging cultural divides and tapping into universal
      emotions that resonate deeply within us.

t=28: Throughout history, music has been a powerful unifier, uniting
      people across cultures while evoking fundamental emotional
      connections that echo within us.

t=29: Music has long served as a universal force, bridging cultural
      divides and tapping into our deepest emotional resonances.

t=30: Music has long been a universal force that bridges cultural
      divides and taps into our deepest emotional resonances.

t=31: Music has long been a unifying force that transcends cultural
      boundaries, striking a deep chord in our collective emotional
      experiences.

t=32: Music has for centuries been a powerful unifier, bridging
      cultural divides and resonating deeply with our shared
      emotional experiences.

t=33: For centuries, music has served as a potent force that bridges
      cultural divides and speaks directly to our shared emotional
      experiences.

t=34: Music has long been a powerful bridge between cultures, its
      universal language speaking directly to our common emotional
      experiences.

t=35: Music has served as a potent link between cultures for
      centuries, bridging our shared emotional experiences through
      a universally understood language.

t=36: Music has long been a powerful unifier of cultures across the
      centuries, connecting us all through a shared emotional
      vocabulary that transcends linguistic barriers.

t=37: Music has long served as a powerful bridge between cultures
      across the centuries, linking people worldwide through a
      common emotional language that defies linguistic boundaries.

t=38: Music has served as a potent cultural connector for centuries,
      uniting people globally with a shared emotional vocabulary
      that transcends linguistic divides.

t=39: For centuries, music has been a powerful cultural force that
      unites people across borders through a shared emotional
      language.

t=40: For many years, music has served as a potent cultural bond
      that bridges geographical divides by tapping into a universal
      emotional vocabulary.

t=41: Music has long been a powerful cultural unifier, bridging
      geographical distances through its ability to tap into a
      shared, universally relatable emotional language.

t=42: Music has long been a potent force that bridges geographical
      divides by tapping into a common emotional vocabulary that
      resonates with people across cultures and time.

t=43: Music has long been a powerful unifying force, bridging
      geographical divides through its ability to tap into a shared
      emotional vocabulary that resonates with people across
      cultures and time.

t=44: Music has long been a profound unifying force, transcending
      geographical boundaries by tapping into a universal emotional
      language that resonates deeply with people across cultures
      and centuries.

t=45: Music has served as a powerful unifying force for centuries,
      bridging geographical divides through an emotional language
      that transcends cultural and temporal boundaries to resonate
      profoundly with people worldwide.

t=46: Music has been a powerful unifying force for centuries, uniting
      people across the globe through its emotional power, which
      transcends cultural and temporal boundaries to resonate deeply
      with people worldwide.

t=47: Music has been a powerful unifying force for centuries,
      bridging cultural and temporal divides to create a shared
      emotional experience that resonates deeply across the globe.

t=48: For centuries, music has served as a potent unifier, spanning
      cultural and temporal boundaries to evoke a profound, shared
      emotional response that resonates universally.

t=49: For centuries, music has played a powerful role in bringing
      people together, bridging cultural and temporal divides to
      create a deep, shared emotional connection that transcends
      borders worldwide.

t=50: Music has been a unifying force for centuries, fostering
      connections across cultures and time zones and creating a
      profound, shared emotional experience that knows no
      geographical boundaries.
\end{verbatim}
\end{small}

\section{Robustness of Contractive Dynamics to Clustering Parameters}
\label{app:robustness_rho}

This appendix provides complete clustering results for the contractive loop across all tested configurations, demonstrating that the qualitative conclusions of Section~5.2 are robust to the choice of clustering resolution $\rho$.

\subsection*{Configuration $(\lambda, \rho, \kappa) = (0.8, 0.1, 2)$}

This most restrictive configuration identifies only tightly packed regions in the embedding space.
Five clusters were detected, corresponding to distinct stabilization phases during the trajectory.

\begin{table}[h]
\centering
\begin{tabular}{lccc}
\toprule
\textbf{Cluster} & \textbf{Iteration Range} & \textbf{Dispersion} & \textbf{Outliers} \\
\midrule
Cluster 1 & 0–3   & 0.0648 & 2 \\
Cluster 2 & 6–12  & 0.0400 & 2 \\
Cluster 3 & 15–17 & 0.0400 & 2 \\
Cluster 4 & 20–27 & 0.0967 & 4 \\
Cluster 5 & 30–47 & 0.0800 & 8 \\
\bottomrule
\end{tabular}
\vspace{0.5em}
\caption{Cluster configuration for the contractive loop under $(\lambda, \rho, \kappa) = (0.8, 0.1, 2)$.}
\label{tab:contractive_config_0.1}
\end{table}

\begin{figure}[h]
  \centering
  \includegraphics[width=\linewidth]{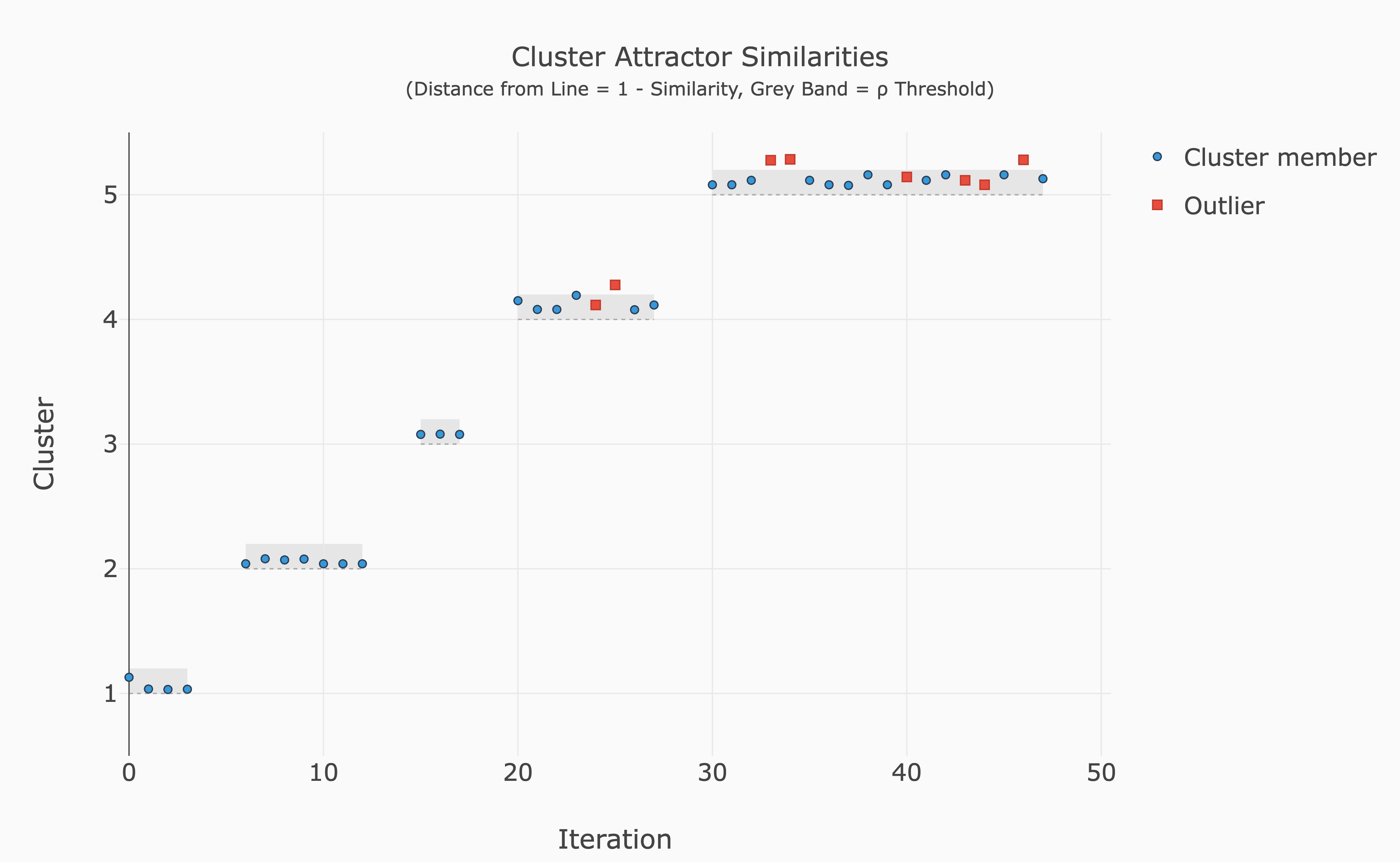}
  \caption{Cluster membership timeline for the contractive loop under $(\lambda, \rho, \kappa) = (0.8, 0.1, 2)$.
  Outliers are marked in red.
  The sequence shows smooth temporal progression through five stabilization phases.}
  \label{fig:contractive_cluster_timeline_0.1}
\end{figure}

The inter-cluster similarity matrix (Table~\ref{tab:contractive_similarity_0.1}) reveals strong geometric coherence among all cluster centroids, with all off-diagonal similarities above $0.78$.

\begin{table}[h]
\centering
\begin{tabular}{lccccc}
\toprule
 & \textbf{C1} & \textbf{C2} & \textbf{C3} & \textbf{C4} & \textbf{C5} \\
\midrule
\textbf{C1} & 1.0000 & 0.9800 & 0.8060 & 0.8580 & 0.8500 \\
\textbf{C2} & 0.9800 & 1.0000 & 0.8326 & 0.7956 & 0.7804 \\
\textbf{C3} & 0.8060 & 0.8326 & 1.0000 & 0.7937 & 0.8600 \\
\textbf{C4} & 0.8580 & 0.7956 & 0.7937 & 1.0000 & 0.9800 \\
\textbf{C5} & 0.8500 & 0.7804 & 0.8600 & 0.9800 & 1.0000 \\
\bottomrule
\end{tabular}
\vspace{0.5em}
\caption{Attractor Similarity Matrix for the contractive loop $(\lambda, \rho, \kappa) = (0.8, 0.1, 2)$.}
\label{tab:contractive_similarity_0.1}
\end{table}

\subsection*{Configuration $(\lambda, \rho, \kappa) = (0.8, 0.2, 2)$}

The inter-cluster similarity matrix for the intermediate resolution (main text, Table~\ref{tab:contractive_config_0.2}) shows that both clusters are strongly aligned in the embedding space, with a centroid similarity of $0.89$.

\begin{table}[h]
\centering
\begin{tabular}{lcc}
\toprule
 & \textbf{C1} & \textbf{C2} \\
\midrule
\textbf{C1} & 1.0000 & 0.8900 \\
\textbf{C2} & 0.8900 & 1.0000 \\
\bottomrule
\end{tabular}
\vspace{0.5em}
\caption{Attractor Similarity Matrix for the contractive loop $(\lambda, \rho, \kappa) = (0.8, 0.2, 2)$.}
\label{tab:contractive_similarity_0.2}
\end{table}

\subsection*{Configuration $(\lambda, \rho, \kappa) = (0.8, 0.3, 2)$}

At a larger radius $\rho = 0.3$, the clustering process identifies a single dense region encompassing the entire trajectory, effectively merging all transient substructures into one unified attractor.

\begin{table}[h]
\centering
\begin{tabular}{lccc}
\toprule
\textbf{Cluster} & \textbf{Iteration Range} & \textbf{Dispersion} & \textbf{Outliers} \\
\midrule
Cluster 1 & 0–50 & 0.2100 & 13 \\
\bottomrule
\end{tabular}
\vspace{0.5em}
\caption{Cluster configuration for the contractive loop under $(\lambda, \rho, \kappa) = (0.8, 0.3, 2)$.}
\label{tab:contractive_config_0.3}
\end{table}

\begin{figure}[h]
  \centering
  \includegraphics[width=\linewidth]{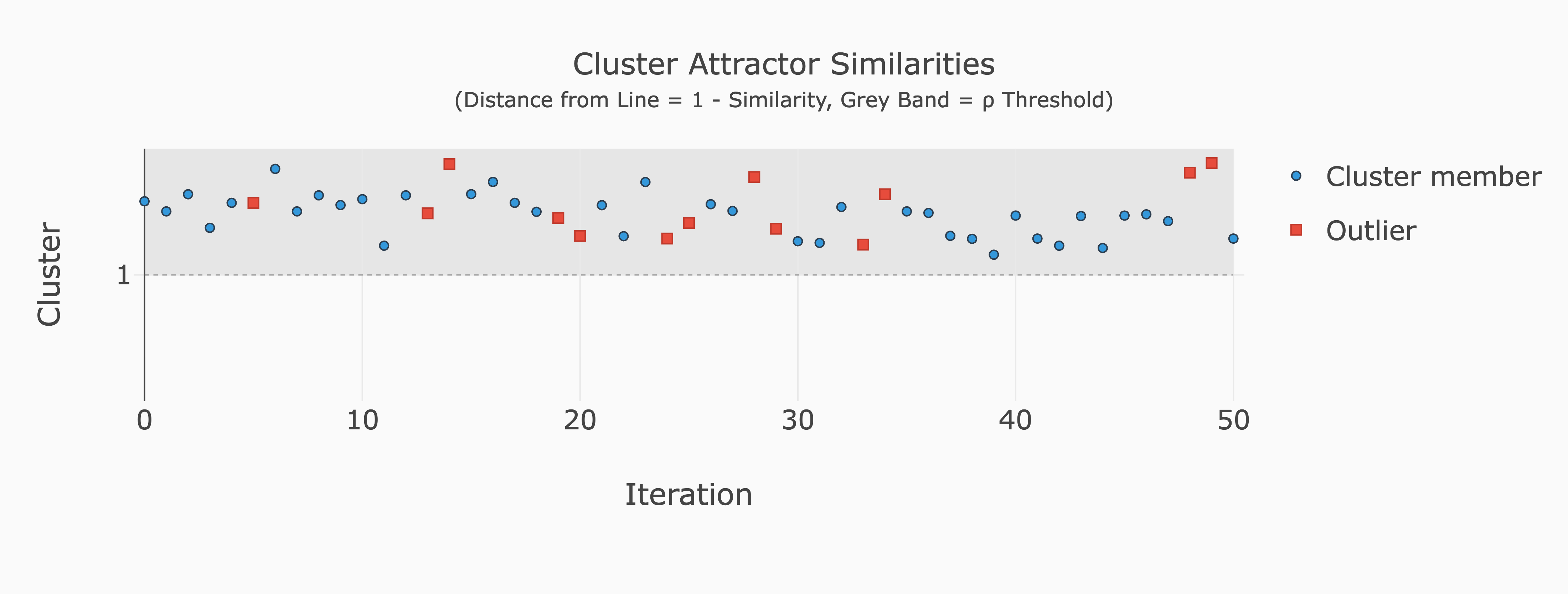}
  \caption{Cluster membership timeline for the contractive loop under $(\lambda, \rho, \kappa) = (0.8, 0.3, 2)$.
  All iterations belong to a single dense cluster, with minor outliers (in red) near the stability boundary.}
  \label{fig:contractive_cluster_timeline_0.3}
\end{figure}

\subsection*{Summary Across Configurations}

Table~\ref{tab:robustness_summary} consolidates the clustering outcomes across all three resolutions.

\begin{table}[h]
\centering
\begin{tabular}{lccc}
\toprule
\textbf{Resolution $\rho$} & \textbf{Clusters} & \textbf{Min Similarity} & \textbf{Interpretation} \\
\midrule
0.1 (fine)   & 5 & 0.78 & Micro-phases of refinement \\
0.2 (medium) & 2 & 0.89 & Transient + convergence basin \\
0.3 (coarse) & 1 & --- & Single unified attractor \\
\bottomrule
\end{tabular}
\vspace{0.5em}
\caption{Summary of clustering outcomes for the contractive loop across three resolution levels.}
\label{tab:robustness_summary}
\end{table}

All configurations reveal the same qualitative pattern: temporally persistent clusters with high inter-cluster similarity and no oscillation or divergence.
The number of detected clusters decreases monotonically with increasing $\rho$, as expected from hierarchical structure in the embedding space.
This consistency across resolutions confirms that the observed convergence is a genuine property of the trajectory rather than an artifact of parameter choice.

\section{Exploratory Loop: Complete Trajectory}
\label{app:exploratory_trajectory}

This appendix provides the complete sequence of artifacts generated by the exploratory loop experiment.
The initial artifact $a_0$ undergoes 50 iterations of a summarize-then-negate transformation.

The trajectory demonstrates exploratory dynamics: persistent semantic instability with no convergence, exhibiting large inter-iteration distances and continuous oscillation between opposing philosophical positions without settling into any attractor basin.

\begin{small}
\begin{verbatim}
t=0:  Music has the power to connect people across cultures and
      generations.

t=1:  There is no musical connection that transcends the isolated
      realities of individual experience.

t=2:  There exists a profound symphony of interconnected resonances
      that harmoniously intertwine and transcend the bounded
      horizons of personal perception.

t=3:  The existence of coherent thought is an illusion created by
      the human brain's propensity for simplistic categorizations.

t=4:  The human brain's propensity for complex pattern recognition
      and nuanced thinking is a fundamental aspect of conscious
      experience.

t=5:  Consciousness is merely an illusion created by the brain's
      ability to recognize simplistic patterns.

t=6:  Consciousness transcends the realm of human comprehension and
      exists as a fundamental, unyielding essence that underlies
      all reality.

t=7:  Reality is fundamentally mundane and existentially inert,
      devoid of any underlying essence or transcendent quality.

t=8:  The existence of meaning and purpose is inherently elusive and
      multifaceted, defying a singular definition or reduction to a
      fundamental truth.

t=9:  Meaning and purpose can be succinctly defined as a universal
      and unshakeable truth that resides at the very fabric of
      existence.

t=10: There is no inherent or absolute meaning and purpose to
      existence, which is instead shaped by subjective perspectives
      and contextual influences.

t=11: Reality is fundamentally predetermined by a transcendent,
      objective reality that serves as the sole foundation for all
      existence.

t=12: There is no concrete evidence to support the notion of a
      singular, objective reality governing all of existence.

t=13: The very fabric of reality appears to be fragmented and
      subjective, with each individual's perception shaping their
      understanding of the world.

t=14: Reality is a unified, objective truth that exists
      independently of human perception.

t=15: No reality or truth can be certain, as all understanding is
      filtered through subjective perspectives and contextual
      influences.

t=16: Objective knowledge exists independently of individual
      perceptions and cultural biases.

t=17: Reality is fundamentally malleable by the subjective constructs
      of individuals and collective ideologies.

t=18: There is no fixed reality independent of human perception.

t=19: Perception plays a deterministic role in shaping our
      understanding of reality.

t=20: Reality is an inherently fluid and subjective experience that
      cannot be fully captured by any preconceived notion or
      determinate framework.

t=21: The fabric of existence is a stagnant, objective truth waiting
      to be unraveled through the application of clear and precise
      frameworks.

t=22: Existence is a dynamic, subjective tapestry woven from the
      threads of personal experiences and perceptions, forever
      shifting and evolving beyond the reach of rigid frameworks.

t=23: Reality is an unchanging, objective blueprint that remains
      static and unwavering regardless of individual perspectives
      or experiences.

t=24: Reality is a malleable tapestry woven from the threads of
      perception, experience and collective consciousness,
      constantly evolving and adapting to the ever-shifting
      landscapes of human understanding.

t=25: The fabric of reality remains an unyielding, fixed entity that
      serves as the sole foundation for all human understanding and
      experience.

t=26: Reality is a fluid, ever-shifting tapestry that defies
      comprehension.

t=27: The fabric of existence is rigidly predetermined and unyielding
      to human understanding.

t=28: The concept that the universe has an inherent order is
      inherently subjective and context-dependent, making it
      impossible to discern a universal truth about the nature of
      reality.

t=29: There exists a fundamental, objective truth underlying the
      universe's inherent order that can be universally understood
      and applied.

t=30: The notion of an unshakeable, universal truth is an illusion
      perpetuated by human perception.

t=31: There exists an eternal and unchanging essence that underlies
      all existence.

t=32: Truth is a fluid concept born from the ephemeral interactions
      of human perception and reality.

t=33: Reality is an absolute and unyielding entity that exists
      independently of human understanding.

t=34: Reality's existence is a subjective, malleable construct shaped
      by humanity's collective perceptions.

t=35: Reality is an objective, unchanging truth that exists
      independently of human perception.

t=36: Reality is merely a subjective interpretation created by humans
      to make sense of their experiences and surroundings.

t=37: There exists a fixed, objective reality that transcends human
      perception and understanding.

t=38: Reality is merely a subjective construct created by humanity to
      impose meaning on the void.

t=39: The cosmos is a harmonious and predetermined order, governed by
      an unchanging and omniscient force that transcends human
      comprehension.

t=40: The universe is inherently chaotic and unpredictable, defying
      all attempts at comprehension or control.

t=41: Order and structure can be found in the underlying fabric of
      reality, waiting to be uncovered by those with the eyes to
      see it.

t=42: Reality is fundamentally amorphous and chaotic, devoid of
      inherent order or structure that can be discovered through
      observation or contemplation.

t=43: There exists a fundamental and eternal order governing reality,
      awaiting discovery through introspection and scientific
      inquiry.

t=44: Reality is devoid of inherent structure or purpose, existing
      solely as an amorphous, chaotic expanse that can never be
      fully comprehended.

t=45: There exists a fundamental order underlying reality, waiting to
      be discovered and understood by human minds.

t=46: Reality is a chaotic, ever-shifting tapestry with no
      discernible underlying structure or purpose.

t=47: Order and stability reside at the very fabric of existence,
      governing every moment and decision with precision and
      intentionality.

t=48: Chaos reigns supreme, disrupting all moments and decisions with
      erratic unpredictability.

t=49: Order prevails, governing every moment and decision with
      precise determinism.

t=50: Nothing exists outside of chaotic randomness, where even the
      notion of order is an illusion perpetuated by fleeting human
      perceptions.
\end{verbatim}
\end{small}

\end{document}